\documentclass{article}

\usepackage{arxiv}

\usepackage[utf8]{inputenc} 
\usepackage[T1]{fontenc}    
\usepackage{hyperref}       
\usepackage{url}            
\usepackage{booktabs}       
\usepackage{amsfonts}       
\usepackage{nicefrac}       
\usepackage{microtype}      
\usepackage{lipsum}
\usepackage{graphicx}
\usepackage{amsmath}
\graphicspath{ {./images/} }
\usepackage[ruled,vlined]{algorithm2e}

\title{MudraGen: Geometrically Supervised Generation of Interacting Two-Hand Mudras for preserving Indian Classical Dance Heritage}

\author{
Jagadish Kashinath Kamble \\
Indian Institute of Technology Kharagpur\\
Kharagpur, West Bengal, INDIA \\
jkkamble@pict.edu \\
\And
Jayanta Mukhopadhyay \\
Indian Institute of Technology Kharagpur\\
Kharagpur, West Bengal, INDIA \\
\And
Debaditya Roy \\
Indian Institute of Technology Kharagpur\\
Kharagpur, West Bengal, INDIA \\
\And
Partha Pratim Das \\
Ashoka University\\
Sonipat, Haryana, INDIA
}

\begin{document}
\maketitle
\begin{abstract}
 
Automatic generation of hand gestures is essential for the  transmission of Indian classical dance and critical for its preservation. Indian classical dance gesture datasets are inherently low-resource, and the canonical Sanskrit definitions of many mudras lack precise textual descriptions, limiting the effectiveness of conventional text-conditioned image generation models. We present \textbf{MudraGen}, a conditional diffusion framework that synthesizes realistic RGB images of \textit{Samyukta Hasta Mudras} -- interactive two-hand gestures from Bharatanatyam (an Indian classical dance form). 
Unlike prior work on simple hand signs or single-hand gestures, MudraGen introduces geometry-aware supervision to capture the precise coordination, anatomical validity, and cultural nuance of interacting hands.
We formulate three geometry-aware objectives: Keypoint Loss for 3D joint alignment, Joint Offset Loss for inter-hand spatial coherence, and Shape Consistency, which serves as an anatomical regularizer by encouraging consistent hand morphology while allowing independent hand poses. Together, these objectives guide the diffusion model toward anatomically plausible and well-coordinated hand configurations, enabling the synthesis of photorealistic and pose-accurate gesture images. Experimental results show that MudraGen surpasses existing state-of-the-art generative approaches in visual realism, anatomical correctness, and preservation of fine hand-pose structure, enabling faithful reproduction of complex Samyukta Hasta mudras. Beyond quantitative gains, its ability to generate culturally grounded and structurally consistent gestures highlights practical applications in cultural preservation and dance education.
\end{abstract}


\section{Introduction}

Hand gestures constitute one of the most expressive forms of non-verbal communication, conveying narrative, emotion, and symbolic meaning across cultures. They play central roles in religious rituals, sign languages, martial traditions, and traditional performing arts, where meaning is communicated through precise anatomical configurations rather than spoken language. Structured gesture systems appear in Balinese dance traditions such as Pendet and Rejang Sari, where hand configurations and movement vocabularies encode cultural and spiritual knowledge \cite{radiusman2021ethnomath,wahyuni2022lexicon}. Among these traditions, Bharatanatyam, one of the oldest surviving classical dance forms of India, employs a codified vocabulary of hand gestures (\textit{mudras}) as a primary medium for storytelling and emotional expression \cite{raju2024pose2gest, jyoti2024gesture}. Grounded in canonical texts such as the \textit{Nāṭya Śāstra} and the \textit{Abhinaya Darpana}, Bharatanatyam defines a rich repertoire of gestures whose meanings are conveyed through carefully prescribed finger articulation, palm orientation, and coordinated body movement. As Bharatanatyam is transmitted primarily through visual demonstration and embodied practice, computational methods for modeling and synthesizing authentic mudras have significant potential for cultural heritage preservation, dance education, and digital archival \cite{unesco,mdpi2021lazgi,abraham2021gesture}.

Among Bharatanatyam hasta mudras (hand gestures), \textit{Samyukta Hasta} (interacting two-hand) gestures present a substantially greater challenge than \textit{Asamyukta Hasta} (single-hand) gestures. While Asamyukta mudras derive their meaning primarily from the articulation of a single hand, the semantics of Samyukta mudras emerge from the coordinated configuration of both hands. Consequently, gesture validity depends not only on the anatomy of each individual hand but also on precise inter-hand relationships, including symmetry or asymmetry, contact, overlap, relative orientation, and spatial alignment. Even minor deviations in finger articulation or hand positioning can alter—or completely invalidate—the intended semantic meaning.

\begin{figure}[h]
    \centering
    \includegraphics[width=\linewidth]{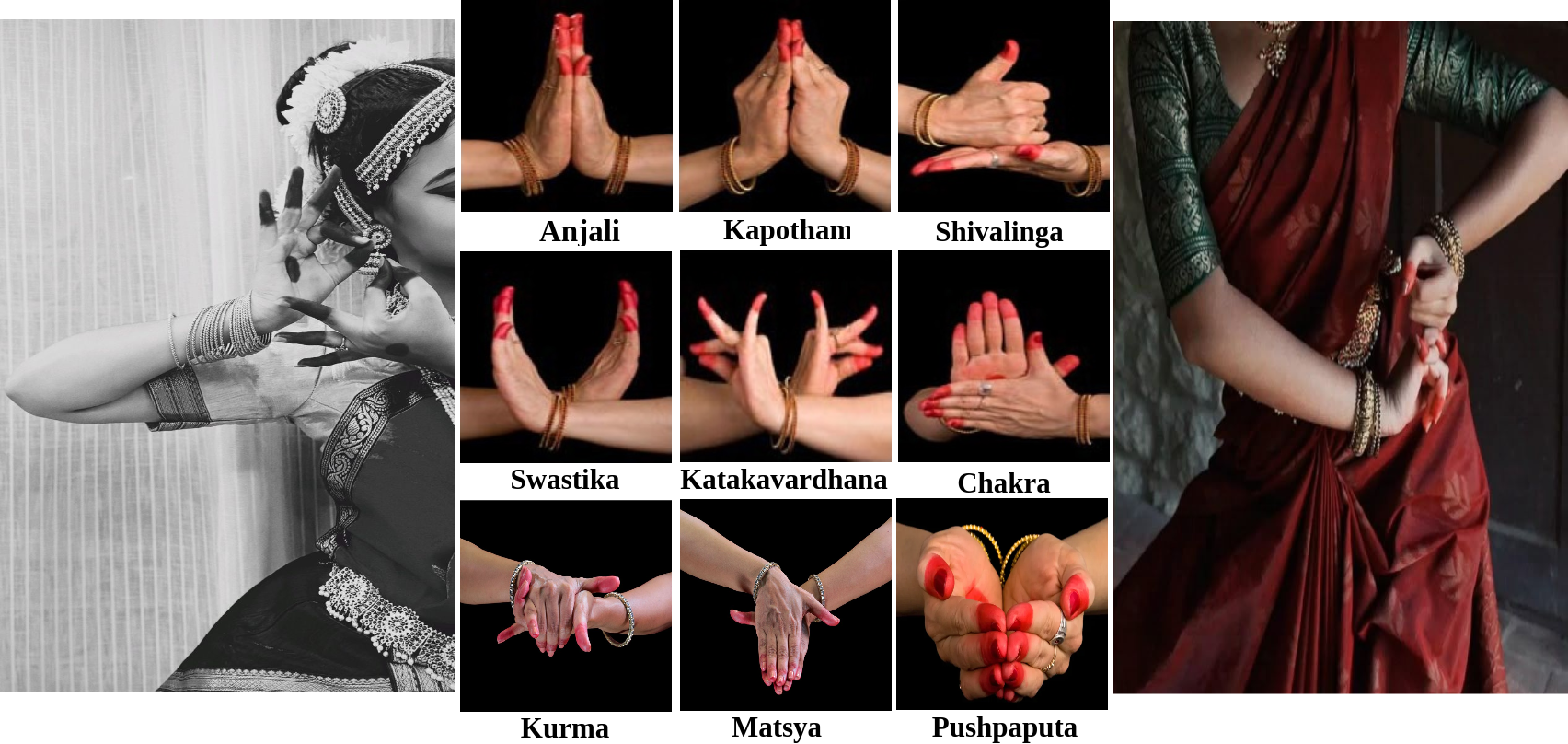}
    \caption{Visual framework highlighting how hasta mudras, particularly Samyukta Hasta, function as essential tools for storytelling through dance. Semantic interpretation depends on precise finger articulation, palm orientation, and coordinated spatial relationships between both hands.}
    \label{fig:samyukta_examples}
\end{figure}

This close relationship between geometry and semantics is illustrated in Fig.~\ref{fig:samyukta_examples}. For example, the gesture \textit{Anjali} conveys reverence through perfect bilateral symmetry and complete palm contact, whereas \textit{Kapotham} requires a hollow enclosure between the palms despite their similar overall appearance. Likewise, \textit{Matsya} and \textit{Kurma} derive their meanings from the coordinated orientation, overlap, and curvature of both hands, while narrative gestures such as \textit{Shivalinga} and \textit{Chakra} require asymmetric hand roles and precise finger interlocking. Across these examples, semantic correctness is inseparable from anatomically accurate articulation and coordinated interacting-hand geometry.

This strong coupling between anatomical structure and semantic meaning makes Samyukta Hasta generation fundamentally different from conventional image synthesis. A generated image may appear visually realistic while still representing an invalid gesture if the underlying finger articulation or spatial relationship between the two hands violates the canonical definition of the mudra. The problem is further complicated by severe self-occlusions and intricate finger interactions, which make reliable modeling of interacting hands substantially more difficult than single-hand generation \cite{zimmermann2017learning}. Three-dimensional hand representations offer a natural mechanism for addressing these challenges by explicitly modeling articulated joints, surface geometry, and inter-hand spatial relationships. Such representations enable geometry-aware supervision through joint alignment, inter-hand consistency, and anatomical regularization, thereby reducing implausible or physically inconsistent hand configurations \cite{zhang2021interactinghands,ren2023dir,tang2018gesturegan,hasson2019towards}.

Recent diffusion-based approaches have demonstrated impressive realism in hand image synthesis. Methods such as Hand100 \cite{hand1000}, HanDiffuser \cite{HanDiffuser}, MUFEEN \cite{fu2024mufeen}, and HandRefiner \cite{lu2024handrefiner} improve generation quality through stronger text alignment, 3D priors, or refinement strategies. However, these methods primarily focus on generic or single-hand generation and rely on natural-language prompts to specify the desired gesture. This assumption is limiting for Bharatanatyam. Canonical mudra definitions originate from Sanskrit treatises and are traditionally transmitted through visual demonstration rather than textual description. Consequently, many structurally distinct mudras lack precise English descriptions. For example, both \textit{Anjali} and \textit{Kapotham} are commonly described as "prayer hands," despite requiring fundamentally different anatomical configurations. Furthermore, Bharatanatyam gesture datasets remain relatively small compared with the large-scale datasets typically used for training modern diffusion models. These characteristics make faithful generation of culturally grounded interacting-hand gestures particularly challenging using existing text-conditioned approaches.

These observations motivate two key design choices in our framework. First, rather than relying on ambiguous textual descriptions, we formulate gesture synthesis as a label-conditioned image generation problem, where each mudra is represented as a distinct visual category. This formulation aligns naturally with the structured taxonomy of Bharatanatyam and avoids the semantic ambiguity introduced by language-based conditioning. Second, because semantic correctness depends on accurate interacting-hand geometry, we incorporate explicit geometry-aware supervision during diffusion training to encourage anatomically plausible articulation and coherent spatial coordination between both hands.

Based on these ideas, we propose MudraGen, a geometry-aware conditional diffusion framework for synthesizing photorealistic RGB images of Bharatanatyam Samyukta Hasta mudras from gesture class labels. During training, a pretrained interacting-hand reconstruction network provides 3D geometric supervision through three complementary objectives: a Keypoint Loss that enforces accurate joint articulation, a Joint Offset Loss that preserves inter-hand spatial coherence, and a Shape Consistency regularizer that encourages anatomically consistent hand morphology. Together, these objectives guide the diffusion model beyond pixel-level realism toward structurally valid and culturally faithful gesture generation. Extensive experiments demonstrate that MudraGen consistently outperforms existing state-of-the-art approaches in both visual realism and structural accuracy. In addition to quantitative evaluation, we validate the generated gestures through multi-view geometric reconstruction and a human expert study conducted with trained Bharatanatyam practitioners, demonstrating that the synthesized mudras preserve both anatomical correctness and cultural authenticity. The key contributions of this work are as follows: \vspace{-0.159cm}
\begin{itemize}
\item We present the first geometry-aware conditional diffusion framework for synthesizing photorealistic RGB images of Bharatanatyam Samyukta Hasta mudras from gesture class labels.

\item We introduce 3D hand-mesh supervision through a pretrained reconstruction network, incorporating Keypoint Loss, Joint Offset Loss, and Shape Consistency regularization to generate anatomically plausible and structurally coherent interacting-hand gestures.

\item We perform extensive quantitative evaluation, multi-view 3D geometric analysis, and a human expert evaluation study with trained Bharatanatyam practitioners, demonstrating that the generated gestures are visually realistic, anatomically accurate, and culturally faithful.

\end{itemize}

\section{Related Work}
\label{sec:Literature_Survey}

\paragraph{Hand Gesture Recognition and Classification}

Hand gesture recognition has been a widely studied topic in computer vision, human-computer interaction (HCI), and sign language translation. Early approaches relied on handcrafted features and rule-based classification using contour, skin color, and motion. With the advent of deep learning, convolutional neural networks (CNNs) and recurrent neural networks (RNNs) became prominent in learning spatiotemporal patterns in hand movements. Notable datasets such as FreiHAND\cite{zimmermann2019freihand}, EgoHands\cite{bambach2015egohands}, HaGRID\cite{kapitanov2023agrid}, and OneHand10K\cite{wang2019maskpose} have enabled the training of robust models for both 2D and 3D hand pose estimation and gesture recognition. Works like Mediapipe\cite{lugaresi2019mediapipe} Hands offers real-time tracking and 21-keypoint estimation, while models like Graphormer\cite{zhao2022graformer}, InterHand2.6M\cite{moon2020interhand2_6m}, and MANO\cite{romero2017mano}, go deeper into 3D mesh and skeletal reconstruction. While these systems exhibit high accuracy for general gesture categories such as "stop", "mute", or "thumbs-up", they frequently lack the fine-grained precision required for semantically rich and culturally specific gestures, particularly those characteristic of Indian classical dance \cite{paul2023fast,raju2024pose2gest}. Research into hand gestures in Indian Classical Dance have primarily concentrated on recognition and classification tasks. Previous studies 
have explored CNN-based models \cite{haridas2022bharatanatyam,kalaimani2024vggmudra}, motion cue fusion \cite{devi2024mcmvbf}, and transfer learning methodologies \cite{parameshwaran2019transfer,naaz2023integrated, nandeppanavar2023hasta} to discern static or dynamic mudras from video or image datasets.

\paragraph{Single hand generation}
Recent studies have gone beyond recognition and explores the  generative modeling for single-hand gestures. Hand1000~\cite{hand1000} explores diffusion-based generation from text using only 1,000 samples, showing the promise of generative models in low-data settings. Its three-stage fusion of gesture recognition, text optimization, and diffusion-based synthesis bridges the gap between language and fine-grained anatomical rendering. HandDiffuser~\cite{HanDiffuser} uses a text-to-image diffusion pipeline to generate realistic single-hand images but struggles with ambiguous articulation and occlusion. AttentionHand~\cite{park2024attentionhand} introduces a text-controllable hand image generation framework that leverages 3D consistency for reconstruction. Earlier work like GestureGAN~\cite{tang2018gesturegan} focuses on gesture-to-gesture translation under controlled conditions but lacks fine-grained anatomical supervision. 

\paragraph{Interacting hand gesture generation}
Generating interacting hand gestures remains a challenging task due to complex articulation, occlusions, \cite{mueller2017real,zimmermann2017learning} and the need for semantic coherence between both hands. FoundHand~\cite{Chen_2024_FoundHand} introduces a 2D keypoint-conditioned diffusion framework trained on a massive dataset of 10M real and synthetic hand images. It supports realistic synthesis of single and dual hand configurations, enabling applications like gesture transfer, domain adaptation, and novel view synthesis. While effective in general gesture modeling, FoundHand primarily focuses on flexible image-level control using 2D keypoints. 
While hand gesture generation has seen growing interest, existing research predominantly focuses on single-hand synthesis, with only a few methods explicitly addressing generic two-hand generation \cite{Chen_2024_FoundHand} .

\paragraph{3D-Aware Gesture Generation}
Recent generative approaches have incorporated 3D hand representations to improve the anatomical and spatial coherence of synthesized gestures. Unlike purely 2D methods, these frameworks leverage mesh-based priors—such as MANO models or implicit surfaces—to guide pose-aware image generation. For example, MUFEEN~\cite{fu2024mufeen} fuses multi-view mesh features for photorealistic single-hand synthesis, while AttentionHand~\cite{park2024attentionhand} bridges text-to-image generation with 3D geometry. Im2Hands~\cite{im2hands2023} reconstructs two-hand 3D meshes from RGB using occupancy fields and keypoint priors, avoiding reliance on parametric models. PoseControl~\cite{lee2025posecontrol} applies RL-tuned ControlNet to better align outputs with keypoint conditions.

\section{Methodology}
\label{sec:method}

Our goal is to synthesize anatomically plausible and semantically accurate \textit{two-hand gestures} (Samyukta Hasta Mudras) from Indian classical dance. Unlike typical text-to-image pipelines, we condition generation solely on gesture class labels. Traditional descriptions of these mudras originate in Sanskrit treatises and pedagogy that emphasize visual demonstration and embodied learning. When expressed through text—particularly in translation—such descriptions lack the spatial specificity required to unambiguously define complex two-hand gestures, making language-driven conditioning unreliable. Modeling Samyukta Hastas as discrete visual classes therefore aligns with the visual transmission practices of Bharatanatyam and enables faithful synthesis of semantically valid gestures. Accordingly, we formulate Samyukta Hasta generation as a label-conditioned generative problem, where each mudra corresponds to a distinct class with well-defined structural constraints.

\subsection{Label-conditioned generation}
Let $\mathcal{Y}$ denote the label space, with each gesture class $y \in \mathcal{Y}$ represented as a one-hot vector. 
We define a conditional generative model $G_\theta$ that produces a realistic RGB image $\hat{x} \in \mathbb{R}^{H \times W \times 3}$ given the label $y$ and random noise $\epsilon \sim \mathcal{N}(0, I)$:

\begin{equation}
\hat{x} = G_\theta(y, \epsilon).
\end{equation}

This design ensures that generation depends only on the class label, allowing us to directly synthesize specific mudras without ambiguous text descriptions. 

For MudraGen, we build on a diffusion-based framework \cite{ho2020ddpm,ho2022classifierfree}, where a clean image $x_0$ is gradually noised through a forward process:

\begin{equation}
q(x_t \mid x_0) = \mathcal{N}\left(x_t; \sqrt{\bar{\alpha}_t},x_0, (1 - \bar{\alpha}_t) I\right),
\end{equation}

with $\bar{\alpha}_t = \prod_{s=1}^{t} \alpha_s$ representing the cumulative noise schedule. During training, the model learns the reverse process by predicting the injected noise $\epsilon$, conditioned on the mudra label $y$:

\begin{equation}
\mathcal{L}_{\text{diff}} = \mathbb{E}_{x_0, y, t, \epsilon} \big[|\epsilon - \epsilon_\theta(x_t, t, y)|_2^2\big],
\end{equation}

where noisy samples are defined as
$x_t = \sqrt{\bar{\alpha}_t} x_0 + \sqrt{1 - \bar{\alpha}_t} \epsilon, \quad \epsilon \sim \mathcal{N}(0, I)$.

At inference, a gesture label $y$ is  sufficient to generate novel two-hand gestures, enabling controlled synthesis of specific mudras.

\begin{figure*}
\centering
\includegraphics[width=\linewidth]{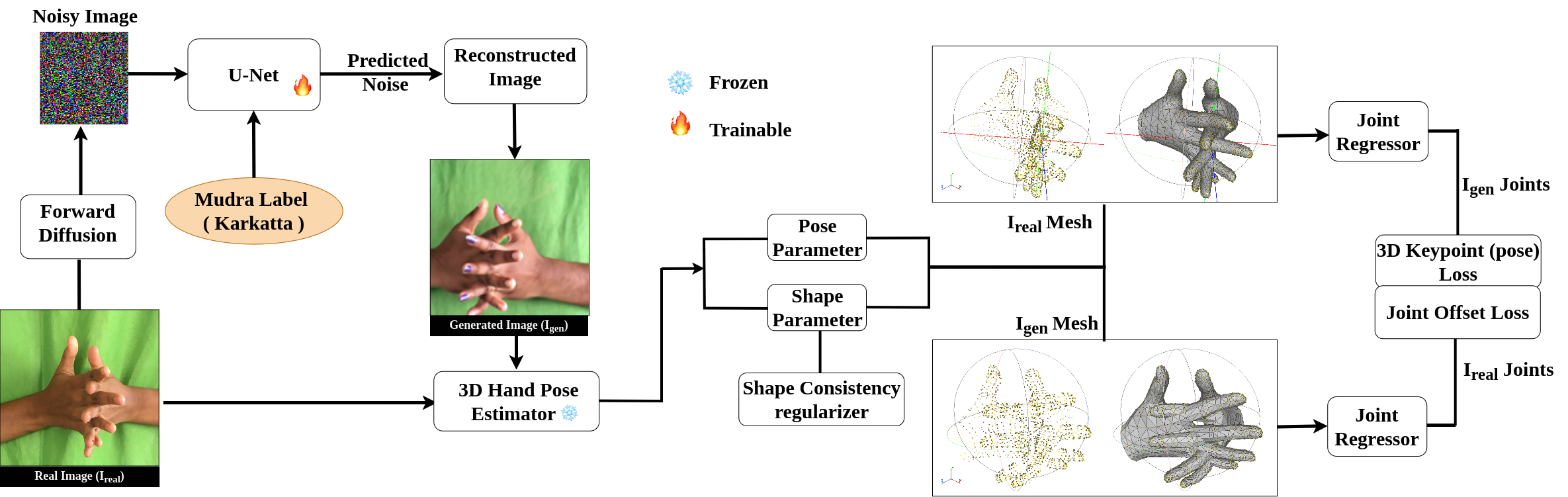}
\caption{Overview of geometry-aware supervision pipeline(training). A class-conditioned U-Net generates symbolic hand gesture images from noise. Both generated ($I_{\text{gen}}$) and real images ($I_{\text{real}}$) are passed through a shared 3D hand mesh regressor model to extract 3D joints and shape parameters. These are used to compute keypoint (pose) loss, joint offset loss, and Shape Consistency regularizer, enabling anatomically accurate and interaction-aware two-hand gesture generation. }
\label{fig:arch_1}
\end{figure*}

\subsection{Image Reconstruction at Diffusion Training}

In Label-conditional diffusion models, the training process extends the standard denoising diffusion framework by incorporating conditioning information, such as class labels, to guide the image generation process. Each training sample consists of a clean image $x_0$ and its corresponding condition $y$ (e.g., class label). The forward diffusion process progressively corrupts $x_0$ with Gaussian noise to produce a noisy version $x_t$ at timestep $t$ (see Eq.~(2)).

During training, the denoising network $\epsilon_\theta(x_t, t, y)$ takes the noisy input $x_t$, the corresponding timestep $t$, and the Label condition $y$ to predict the noise component $\epsilon$ that is added during the forward diffusion process. The conditional diffusion loss is formulated in Eq.~(3),
which trains the model to estimate the true noise for each Label-conditioned instance.

Although this objective allows the model to learn a conditional denoising process that approximates the Label-specific data distribution $p(x_0 | y)$, it does not explicitly ensure geometric or perceptual alignment between the reconstructed and original images. To incorporate such alignment, we compute the reconstructed (generated) image $\hat{x}_0$ from the network’s predicted noise as:
\begin{equation}
\hat{x}_0 = 
\frac{x_t - \sqrt{1 - \bar{\alpha}_t}\,
\epsilon_\theta(x_t, t, y)}{\sqrt{\bar{\alpha}_t}}.
\label{eq:cond_denoised_image}
\end{equation}
Both real and generated images are passed through pretrained 2D and 3D pose estimators to extract keypoints, joint angles, and meshes. Using these structured inputs, we formulate the following losses:
\begin{itemize}
    \item \textbf{Keypoint Loss} measures the Euclidean distance between predicted and ground-truth hand joint coordinates to ensure anatomically accurate joint placement.
    \item \textbf{Joint Offset Loss} penalizes errors in the relative displacement between neighboring joints, encouraging consistent local bone lengths and physically valid hand articulation.
    \item \textbf{Shape Consistency regularizer} enforces structural regularity by ensuring that the predicted hand shape preserves global geometric relationships and remains consistent across viewpoints or poses.
\end{itemize}

\subsection{Geometric Supervision}
\label{sec:losses}
While the diffusion process captures appearance-level fidelity, ensuring structural validity of interacting hands remains a major challenge. 
As illustrated in Fig.~\ref{fig:arch_1}, we introduce structured geometric supervison into our diffusion-based generative pipeline, we convert an RGB image of a Samyukta Hasta Mudra (two interacting hands) into a 3D representation comprising mesh vertices and joint keypoints. 
While RGB images capture appearance-level details, they lack the structural cues needed for pose-level supervision. 
We bridge this gap by introducing structural guidance into the generative process, we leverage MANO (Model with Articulated and Non-rigid Deformations) ~\cite{romero2017mano,zhang2021interactinghands} model which is a parametric 3D hand model that reconstructs a realistic hand mesh from a low-dimensional set of pose $(\boldsymbol{\theta})$ and shape parameters $(\boldsymbol{\beta})$. 
The MANO model takes these parameters and generates a posed 3D hand mesh through several stages. First, it starts with a mean hand mesh template $\mathbf{T}_0$ consisting of 778 vertices. The shape parameters $\boldsymbol{\beta}$ deform this template via a set of learned shape blend shapes $\mathbf{S}_i$ to produce a person-specific hand mesh. This deformation is computed as:
\[
\mathbf{T}_{\text{shape}} = \mathbf{T}_0 + \sum_{i=1}^{10} \beta_i \mathbf{S}_i,
\]
resulting in a mesh that reflects the underlying hand identity. Next, the pose parameters $\boldsymbol{\theta}$ introduce pose-dependent deformations using pose blend shapes $\mathbf{P}_j$, which capture changes in hand surface geometry due to joint bending (e.g., skin folds, muscle bulges). These are applied to the shape-deformed mesh to yield a pose-adjusted mesh, $\mathbf{T}_{\text{pose}}$, that reflects both shape and joint articulation. To articulate the mesh based on joint rotations, the model uses Linear Blend Skinning (LBS), a common technique in computer graphics. Each vertex in the hand mesh is associated with a set of joints through a precomputed skinning weight matrix $W \in \mathbb{R}^{778 \times 16}$, which determines how much each joint influences a given vertex. The final posed mesh $\mathbf{V}$ is computed by transforming each vertex via a weighted combination of the global transformations of the influencing joints: \[
\mathbf{V}_i = \sum_{k=1}^{16} w_{ik} \cdot \mathbf{G}_k \cdot \mathbf{T}_{\text{pose}, i},
\] where $\mathbf{G}_k$ is the transformation matrix for joint $k$. This produces the final 3D hand mesh in a canonical or global coordinate frame. In addition to the mesh, the MANO model also predicts 3D joint locations $\mathbf{J} \in \mathbb{R}^{21 \times 3}$, which are obtained by applying a joint regressor — a fixed matrix $\mathbf{W}_{\text{joint}} \in \mathbb{R}^{21 \times 778}$ — that maps the mesh vertices to joint positions. This enables the model to output both the mesh and the skeletal joint configuration. The final outputs are the vertex positions $\mathbf{V}$ and joint positions $\mathbf{J}$, both in 3D space, serve as the basis for our geometric loss terms—including keypoint accuracy, inter-hand offset consistency, and shape symmetry. 

\subsubsection{Keypoint Loss}
The first challenge is ensuring that each hand is articulated in a biologically plausible way. In Bharatanatyam mudras, even a slight deviation in finger curvature (e.g., straight vs. bent index finger) can change the meaning of a gesture. Purely image-based supervision does not capture these subtleties. The Keypoint Loss directly penalizes discrepancies between predicted and ground-truth 3D joint locations, enforcing anatomically precise articulation of all 42 joints (21 per hand):
\begin{equation}
\mathcal{L}_{\text{kp}} = \frac{1}{2K} \sum_{i=1}^{K} \left\| \hat{\mathbf{J}}_{\text{right}, i} - \mathbf{J}_{\text{right}, i} \right\|_2^2 + \left\| \hat{\mathbf{J}}_{\text{left}, i} - \mathbf{J}_{\text{left}, i} \right\|_2^2
\end{equation}
Here, $K$ is the number of joints per hand, \( \hat{\mathbf{J}}_{\text{right}, i} \in \mathbb{R}^3 \) is the predicted 3D coordinate of the \( i^{\text{th}} \) joint of the right hand, \( \mathbf{J}_{\text{right}, i} \in \mathbb{R}^3 \) is the ground-truth 3D coordinate of the \( i^{\text{th}} \) joint of the right hand, \( \hat{\mathbf{J}}_{\text{left}, i} \in \mathbb{R}^3 \) and \( \mathbf{J}_{\text{left}, i} \in \mathbb{R}^3 \) are defined analogously for the left hand. This term ensures that each hand follows plausible and precise anatomical articulation independently, which is critical for capturing the fine-grained structure of complex gestures.

\subsubsection{Joint Offset Loss}
Two-hand gestures add another layer of complexity: both hands must coordinate spatially to form a single symbolic unit.
While the keypoint loss ensures individual hand articulation plausibility, it does not explicitly account for the spatial relationship between the two hands. 
To address this, we introduce a joint offset loss that supervises the relative positioning between corresponding joints of the left and right hands. Specifically, we compare the inter-hand vector offsets between predicted and ground-truth joints:
\begin{equation}
\mathcal{L}_{\text{offset}} = \frac{1}{K} \sum_{i=1}^{K} \left\| (\hat{\mathbf{J}}_{\text{right}, i} - \hat{\mathbf{J}}_{\text{left}, i}) - (\mathbf{J}_{\text{right}, i} - \mathbf{J}_{\text{left}, i}) \right\|_2^2.
\end{equation}
This loss captures the relational dynamics between the two hands—such as symmetry, mirroring, and contact-based interactions—which are essential in the case of Samyukta Mudras where both hands collaboratively convey a single semantic gesture. By enforcing consistency in inter-hand joint distances, the model learns to generate coordinated and interaction-aware hand poses.

\subsubsection{Shape Consistency regularizer}
Finally, the overall hand shape must remain consistent across both hands. Without explicit constraints, generative models may produce asymmetric or malformed hands—for example, one hand appearing smaller, thinner, or anatomically distorted compared to the other. Such artifacts are especially problematic in culturally codified gestures, where symmetry conveys harmony and balance.
The Shape Consistency regularizer minimizes the $L_2$ distance between the shape parameters $\hat{\boldsymbol{\beta}}$ estimated for each hand of generated images:
\begin{equation}
\mathcal{L}_{\text{shape}} = \left\| \hat{\boldsymbol{\beta}}_{\text{right}} - \hat{\boldsymbol{\beta}}_{\text{left}} \right\|_2^2
\end{equation}
This regularization is especially useful in cases of partial occlusion, projection distortion, or when one hand dominates the visual field. 

The objective of the proposed geometric guidance is not to recover exact ground-truth three-dimensional hand geometry, but rather to impose anatomically meaningful structural constraints during image generation. Specifically, the pose, joint, and shape objectives encourage the generated samples to preserve realistic finger articulation, relative spatial configuration, and consistent interacting-hand geometry. Rather than computing these objectives on the noisy diffusion state, the proposed framework first reconstructs the clean image estimate ($\hat{x}_0$) and extracts geometric representations using the pretrained InterShape network. Since ($\hat{x}_0$) closely resembles a clean RGB image, the reconstruction network operates within its intended input distribution, producing reliable estimates of hand pose and shape. These geometry-aware objectives therefore act as structural regularizers that complement the diffusion denoising objective by discouraging anatomically implausible hand configurations while preserving the visual realism of the generated mudras. Importantly, the proposed framework does not require ground-truth 3D annotations; instead, the pretrained reconstruction network provides a consistent source of geometric guidance that improves the structural plausibility of interacting-hand generation.
\paragraph{Full Training Objective} Our overall training objective combines the standard diffusion loss with the above geometric supervision components. The total loss is defined as:

\begin{equation}
\mathcal{L}_{\text{total}} = \mathcal{L}_{\text{diff}} + \lambda_{\text{kp}} \mathcal{L}_{\text{kp}} + \lambda_{\text{offset}} \mathcal{L}_{\text{offset}} + \lambda_{\text{shape}} \mathcal{L}_{\text{shape}}
\label{eq:total_loss}
\end{equation}

where $\lambda_{\text{kp}}, \lambda_{\text{offset}}, \lambda_{\text{shape}}$ are scalar hyperparameters that balance the contribution of each loss term. The diffusion loss $\mathcal{L}_{\text{diff}}$ governs the core diffusion training, while the additional 3D geometric losses act as soft constraints, guiding the model to synthesize hand images that are not only visually realistic but also structurally valid and culturally faithful.

\begin{algorithm}[t]
\caption{Training Procedure for Geometry-Aware Mudra Generation (MudraGen)  }
\label{alg:mudragen}
\KwIn{Ground-truth image $x_0$, mudra label $y$, diffusion model $\epsilon_\theta$, pretrained InterShape\cite{romero2017mano,zhang2021interactinghands} network}
\KwOut{Updated diffusion model parameters $\theta$}

Sample a diffusion timestep $t \sim \mathcal{U}(1,T)$\;

Sample Gaussian noise $\epsilon \sim \mathcal{N}(0,I)$\;

Generate noisy image:
\[
x_t=\sqrt{\bar{\alpha}_t}x_0+\sqrt{1-\bar{\alpha}_t}\epsilon
\]

Predict noise:
\[
\hat{\epsilon}=\epsilon_\theta(x_t,t,y)
\]

Compute diffusion loss:
\[
\mathcal{L}_{diff}=||\epsilon-\hat{\epsilon}||_2^2
\]

Reconstruct the clean image:
\[
\hat{x}_0=
\frac{x_t-\sqrt{1-\bar{\alpha}_t}\hat{\epsilon}}
{\sqrt{\bar{\alpha}_t}}
\]

Resize $\hat{x}_0$ and $x_0$ to $256\times256$\;

Extract MANO pose, shape, mesh vertices and joints using the pretrained InterShape network\;

Compute:
\begin{itemize}
\item Keypoint loss $\mathcal{L}_{kp}$
\item Joint Offset loss $\mathcal{L}_{offset}$
\item Shape Consistency regularizer $\mathcal{L}_{shape}$
\end{itemize}

Compute total loss:
\begin{equation}
\mathcal{L}_{\text{total}} = \mathcal{L}_{\text{diff}} + \lambda_{\text{kp}} \mathcal{L}_{\text{kp}} + \lambda_{\text{offset}} \mathcal{L}_{\text{offset}} + \lambda_{\text{shape}} \mathcal{L}_{\text{shape}}
\end{equation}

Update diffusion model parameters using backpropagation.

\end{algorithm}
\vspace{-0.3cm}

\begin{figure}
\centering
\includegraphics[width=0.8\linewidth]{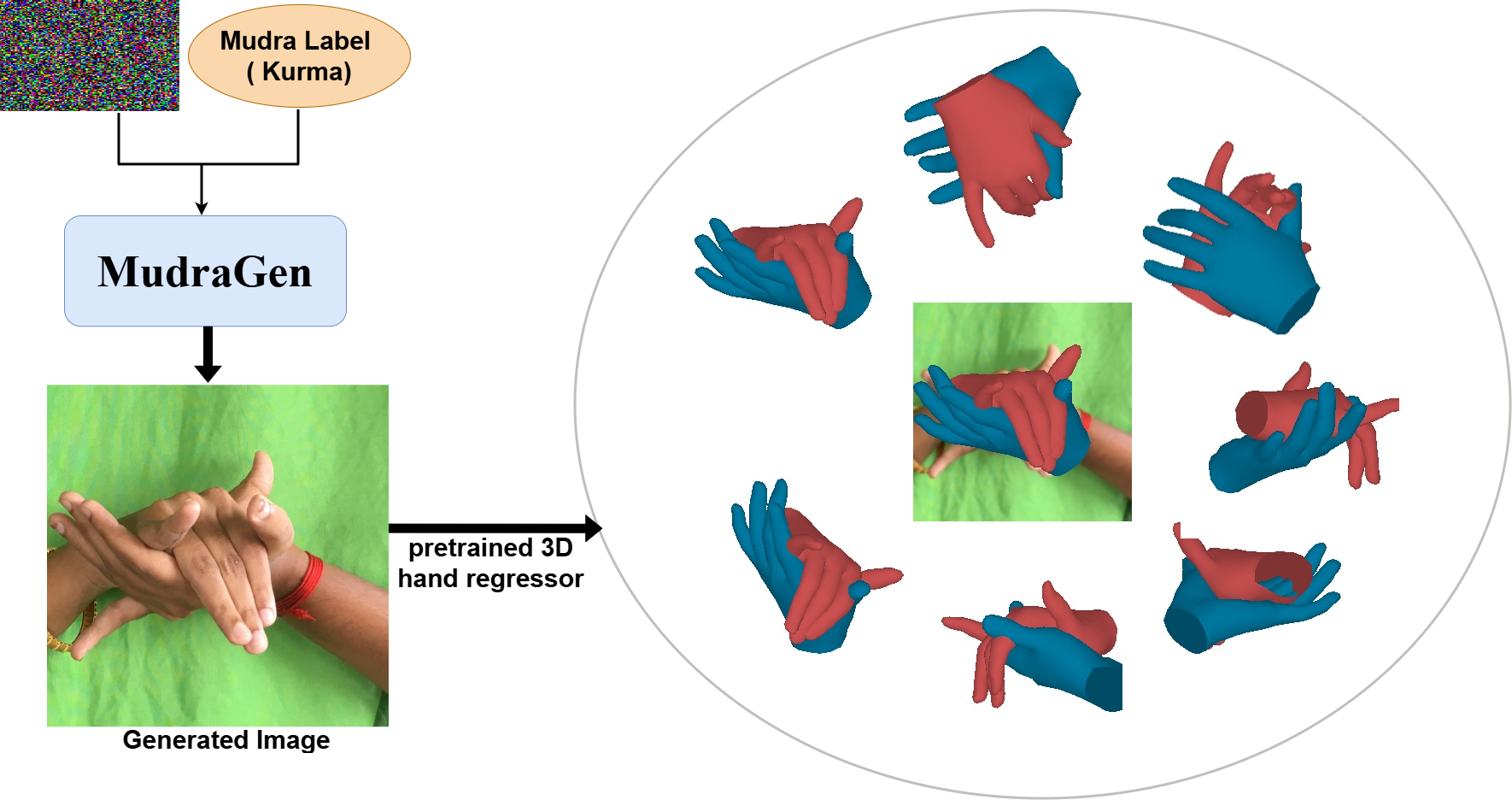}
\caption{Inference Pipeline:End-to-End Generation of Samyukta Mudras with 3D Multi-View Visualization. MudraGen takes noise and a class label as input and generates a gesture image using conditional diffusion. A pretrained 3D hand regressor ~\cite{Li2022intaghand} then converts the generated image into a 3D mesh, which is visualized from six different viewpoints.}
\label{fig:inference_pipeline}
\end{figure}

\paragraph{Inference Pipeline:} During Inference, MudraGen synthesizes a photo-realistic RGB image given a random noise vector and a conditioning Hasta Mudra class label. To analyze and validate the generated gestures geometrically, we pass the generated mudra image through a pretrained 3D hand reconstruction network (IntaHand~\cite{Li2022intaghand}) to extract dense mesh and joint-level representations. As shown in Fig.~\ref{fig:inference_pipeline}, the resulting 3D mesh—consisting of both hands in interacting configuration—is rendered from multiple views, including front, back, left, right, top, bottom, and custom camera angles. This multi-view rendering enables thorough structural inspection, supports cultural gesture understanding, and demonstrates the 3D consistency of the generated output.

The complete training procedure of the proposed geometry-aware conditional diffusion framework is summarized in Algorithm~\ref{alg:mudragen}.

\section{Experimental Results and Performance Evaluation}
\label{sec:experiments}

\subsection{Dataset}
\label{sec:dataset}
We evaluate the proposed framework on the publicly available Bharatanatyam Samyukta Hasta Mudra dataset introduced by \cite{raj2023optimal}. The dataset comprises 13,035 RGB images representing 21 categories of double-hand (Samyukta Hasta) mudras defined according to the classical \textit{Natyashastra}, with approximately 600 images available for each gesture class. All gestures were performed by professionally trained Bharatanatyam dancers with more than five years of formal training, thereby ensuring the semantic correctness and anatomical authenticity of the captured hand configurations.

Unlike large-scale image generation datasets containing millions of images, Bharatanatyam gesture datasets remain inherently low-resource, exhibiting limited sample diversity with respect to performers, viewpoints, illumination conditions, and hand articulations. These characteristics make learning the underlying distribution of anatomically plausible interacting-hand gestures considerably more challenging. Consequently, this dataset provides a realistic benchmark for evaluating conditional generative models under low-resource cultural heritage settings.

\subsection{Implementation Details and Training Protocol}

The proposed framework was implemented in PyTorch and trained on an NVIDIA H100 GPU. Unless otherwise specified, all experiments were performed using RGB images resized to 128×128. The generative backbone is a class-conditional DDPM employing the ContextUNet architecture.

\paragraph{Dataset Preparation.}

The proposed framework is trained using the Bharatanatyam Samyukta Hasta dataset described in Section~\ref{sec:dataset}. To maximize the diversity of samples available for learning the underlying gesture distribution, the dataset is randomly partitioned into 90\% training and 10\% validation subsets using the \texttt{random\_split} utility provided by PyTorch. The training subset is used to optimize the diffusion model, whereas the validation subset is employed to monitor convergence and select the best-performing model checkpoint. To ensure reproducibility across experimental runs, the dataset partition is generated using a fixed random seed of 42.

Since Bharatanatyam gesture datasets are inherently low-resource compared with the large-scale datasets commonly used for training diffusion models, preserving a sufficiently large training set is essential for accurately learning the underlying data distribution. Consequently, rather than introducing a separate held-out test partition, the majority of available samples are utilized for model optimization while the validation subset is used for convergence monitoring and model selection. This strategy maximizes the diversity of training examples without compromising reproducibility and follows a practical protocol for data-constrained generative modeling.

Prior to training, all RGB images are resized to $128\times128$ pixels, normalized to the range $[-1,1]$, and paired with their corresponding Samyukta Hasta labels, which serve as the conditioning signal throughout the reverse diffusion process.

\paragraph{Diffusion Configuration.}
The forward diffusion process consists of $T=400$ timesteps using a linear variance schedule $\beta_1=10^{-4}$, $\beta_T=2\times10^{-2}$. During each training iteration, a timestep $ t\sim\mathcal U(1,T) $ is uniformly sampled, and Gaussian noise $\epsilon\sim\mathcal N(0,I)$ is added to the clean image according to the standard DDPM forward process. The denoising network is trained to predict the injected noise using the conventional DDPM objective.

Unlike the original DDPM implementation that employs $T=1000$ diffusion steps, we adopt $T=400$ throughout all experiments. Preliminary experiments indicated that increasing the diffusion trajectory beyond 400 steps provided only marginal improvements (refer Table~\ref{tab:timestep_analysis}) in generation quality while substantially increasing computational cost. Since the proposed framework operates on relatively low-resolution ($128\times128$) gesture images with limited visual complexity, $T=400$ provides an effective trade-off between synthesis quality and computational efficiency.

\paragraph{Network Architecture.}
The proposed model employs a Context U-Net architecture, an extension of the conventional U-Net that incorporates auxiliary contextual information to guide the prediction process. 

The denoising network adopts the ContextUNet architecture, which extends the conventional U-Net by incorporating diffusion timestep embeddings together with gesture-label embeddings for conditional image generation. The encoder consists of residual convolutional blocks arranged in three hierarchical downsampling stages, whereas the decoder reconstructs high-resolution image details using transposed convolutions and skip connections. Temporal embeddings encode the diffusion timestep, while learnable gesture-label embeddings are injected through feature modulation to guide the denoising process. This design enables the network to simultaneously capture global hand configurations and fine-grained finger articulations required for anatomically plausible Samyukta Hasta generation.

\paragraph{Optimization.}
Training is performed using the Adam optimizer with an initial learning rate of $ 1\times10^{-4}$. The batch size is set to $20$, and the model is trained for $200$ epochs. The learning rate is linearly decayed throughout training according to
\[
\eta_e=\eta_0\left(1-\frac{e}{E}\right),
\]
where $\eta_0$ denotes the initial learning rate, $e$ is the current epoch, and $E$ denotes the total number of epochs. Model parameters were updated after every mini-batch using standard backpropagation.

\paragraph{Geometry-aware Supervision.}

The model is optimized using the overall objective defined in Eq.~(\ref{eq:total_loss}), which combines the diffusion denoising loss with the proposed keypoint, joint-offset, and shape consistency terms. During training, the weighting coefficients are empirically fixed as
\[
\lambda_{\text{kp}} = \lambda_{\text{offset}} = \lambda_{\text{shape}} = 1.
\]

Prior to optimization, the geometric quantities are normalized, resulting in comparable numerical magnitudes across all auxiliary objectives. Consequently, the auxiliary objectives are assigned equal weights ($\lambda_{\text{kp}} = \lambda_{\text{offset}} = \lambda_{\text{shape}} = 1$), which provides stable optimization without introducing additional hyperparameters while allowing the diffusion denoising objective to remain the primary learning signal. The contribution of each geometric objective is further validated through the ablation study presented in Table~\ref{tab:ablation-loss}, where removing any individual loss consistently degrades generation quality.
\paragraph{Geometric Loss Computation}

For every training iteration, a diffusion timestep is first sampled and the noisy image $x_t$ is generated. The diffusion network predicts the injected noise $\epsilon_\theta(x_t,t,y)$, from which the clean image estimate $\hat{x}_0$ is analytically reconstructed using Eq.~(\ref{eq:cond_denoised_image}). Rather than computing geometric supervision directly on the noisy intermediate sample $x_t$, all geometric objectives are evaluated using the reconstructed clean image $\hat{x}_0$. This strategy allows the pretrained InterShape network to operate on visually meaningful hand images that remain close to its original training distribution, thereby improving optimization stability and preventing unreliable predictions from heavily corrupted diffusion samples.

Since the diffusion model generates images at $128\times128$, whereas the pretrained InterShape network expects $256\times256$ inputs, both the reconstructed image $\hat{x}_0$ and the corresponding ground-truth image are resized to $256\times256$, normalized to the range $[-0.5,0.5]$, and processed by InterShape. This resizing is performed exclusively for geometric supervision and does not modify the operating resolution of the diffusion model.

InterShape extracts deep visual features using a ResNet-50 backbone before regressing MANO pose parameters $\boldsymbol{\theta}\in\mathbb{R}^{48}$ and shape parameters $\boldsymbol{\beta}\in\mathbb{R}^{10}$ for both hands. These parameters are subsequently passed through the differentiable MANO layer to recover the complete 3D hand meshes and corresponding three-dimensional joint locations. The resulting representations are used to compute the proposed keypoint loss, joint-offset loss, and shape consistency regularizer. Since the complete reconstruction pipeline remains differentiable, gradients from the geometric objectives propagate through the reconstruction network and ultimately update the parameters of the diffusion model in an end-to-end manner. The proposed geometric supervision is applied only during training. It introduces a modest training overhead due to two additional forward passes through the frozen InterShape network for the generated and ground-truth images and is used solely for feature extraction. Since the InterShape network is discarded after training, the proposed method incurs no additional inference-time overhead.

The shape consistency term functions as an anatomical regularizer rather than direct supervision. Since MANO shape parameters describe intrinsic hand morphology rather than articulation, the regularizer encourages anatomically consistent hand geometry while allowing both hands to assume independent pose configurations. Although this assumption is appropriate for most Samyukta Hastas, highly asymmetric gestures may benefit from gesture-dependent adaptive weighting of the shape regularizer, which remains an interesting direction for future investigation.

\paragraph{Reconstruction of image during loss computation.} Keypoint, offset, and shape losses require the reconstruction of the image  $\hat{x}_0$. We show that it can be performed at negligible computational cost making the losses computationally feasible.
The reconstruction of the image in conditional diffusion models is given by the closed-form expression in Eq.~(\ref{eq:cond_denoised_image}).
This computation introduces \emph{no additional neural network layers} and consists of: one element-wise multiplication, one element-wise subtraction, one element-wise division. All tensors involved in \eqref{eq:cond_denoised_image} already reside in GPU memory and participate in the diffusion loss computation. Because GPUs execute such element-wise operations in parallel, the computational overhead of constructing the image ($\hat{x}_0$) during training is negligible relative to a single UNet forward pass. We validate this empirically by comparing: UNet-only forward pass and UNet + $\hat{x}_0$ reconstruction, consisting of the forward pass and the analytic reconstruction in \eqref{eq:cond_denoised_image}.

\begin{table}[h!]
\centering
\caption{Runtime comparison between UNet with and without image ($\hat{x}_0$) reconstruction. Overhead is negligible.}
\vspace{4pt}
\label{tab:timing}
\begin{tabular}{l c}
\toprule
\textbf{Operation} & \textbf{Mean Time (ms)} \\
\midrule
UNet only & $9.69$ \\
UNet + $\hat{x}_0$ reconstruction & $9.78$ \\
\midrule
\textbf{Overhead} & $\mathbf{0.09}$ \\
\bottomrule
\end{tabular}
\end{table}

As shown in Table~\ref{tab:timing}, the additional reconstruction step introduces an overhead of 0.09ms which is less than $1\%$ of the UNet forward pass time.  Therefore, $\hat{x}_0$ can be computed at every timestep during diffusion training without affecting throughput or training efficiency.

\paragraph{Inference.}
During inference, image synthesis begins from Gaussian noise and proceeds through the complete 400-step reverse diffusion process conditioned on the desired Samyukta Hasta label. Classifier-free guidance is applied with a guidance scale of 2.0 to improve semantic consistency between the generated image and the conditioning label. The pretrained InterShape network is not required during inference, as it serves exclusively as a differentiable geometry-supervision module during training.

Following image generation, the synthesized RGB image is processed using the pretrained IntaHand~\cite{Li2022intaghand} model solely for qualitative 3D reconstruction and multi-view visualization, as described in Section~\ref{sec:method}. IntaHand is not involved in the optimization process, does not influence the generated images, and is not used for quantitative evaluation. Instead, it serves as an independent post-generation reconstruction framework for qualitative geometric analysis, reducing reliance on the supervision network used during training while providing an additional assessment of the anatomical plausibility of the generated interacting-hand gestures.
\begin{table}[h!]
\centering
\caption{{Effect of Diffusion Timesteps}}
\vspace{4pt}
\label{tab:timestep_analysis}

\begin{tabular}{l c c c c}
\toprule

\textbf{Diffusion Steps} & \textbf{FID($\downarrow$)} & \textbf{KID($\downarrow$)} & \textbf{LPIPS ($\downarrow$)} & \textbf{MS-SSIM($\uparrow$)}  \\ \midrule
200 & 61.6932 & 0.0256& 0.2890& 0.5723\\
400 & 51.9126 & 0.0132& 0.1640 & 0.7534 \\
600 & 51.9123 & 0.0130& 0.1637& 0.7557\\
1000 & 51.9121 & 0.0131& 0.1635& 0.7561\\

\bottomrule
\end{tabular}
\end{table}

\paragraph{Design Rationale.}
The proposed design choices are motivated by both the characteristics of the Bharatanatyam gesture generation task and computational efficiency. Diffusion models are selected because they effectively learn complex multimodal image distributions while producing stable training dynamics and high-fidelity image synthesis, making them well suited for generating diverse interacting-hand configurations from discrete gesture labels. The ContextUNet architecture is adopted to incorporate gesture-conditioning information through contextual embeddings while simultaneously capturing global inter-hand relationships and fine-grained finger articulations using its hierarchical encoder--decoder structure. Classifier-free guidance enables controllable conditional generation without requiring additional conditioning networks or auxiliary optimization objectives, thereby maintaining a simple training framework. Geometry-aware supervision is introduced to explicitly constrain the generated images using three-dimensional hand structure, encouraging anatomically plausible finger configurations and consistent hand interactions that cannot be sufficiently enforced by the diffusion objective alone. Furthermore, the diffusion model operates at a resolution of $128\times128$ to balance image quality and computational efficiency, whereas geometric supervision is performed at $256\times256$ to match the input resolution expected by the pretrained InterShape reconstruction network. Finally, $T=400$ diffusion steps are adopted as they provide an effective trade-off between generation quality and computational cost for the relatively low-resolution gesture images considered in this work.

\subsection{Outcomes}

\paragraph{Baseline Methods}

Since no publicly available generative framework has been specifically developed for interacting Samyukta Hasta Mudra generation, we compare the proposed framework against representative conditional image generation methods that constitute the closest available baselines.

\text{Stable Diffusion \cite{rombach2022high}} is included as a strong general-purpose image generation baseline. Recent latent diffusion models have demonstrated impressive image synthesis capabilities through large-scale language-image pretraining. However, Bharatanatyam mudras are defined using canonical Sanskrit terminology whose semantic meaning is deeply rooted in cultural and performative traditions and often lacks precise textual descriptions. Consequently, although Stable Diffusion is capable of generating visually realistic hand images, it does not explicitly model the fine-grained geometric constraints required for anatomically accurate Samyukta Hasta generation. For fair comparison, Stable Diffusion is fine-tuned using the Bharatanatyam Samyukta Hasta dataset under the same experimental protocol adopted for the proposed framework.

\text{Hand1000 \cite{zhang2024hand1000}} is selected as a representative hand-specific generative framework. Originally developed for conditional hand image generation using learned hand representations, the model primarily focuses on general hand appearance rather than interacting-hand geometry. For this work, the framework is adapted and fine-tuned on the Bharatanatyam dataset to synthesize interacting two-hand gestures. Unlike the proposed method, Hand1000 does not incorporate explicit three-dimensional geometric supervision during optimization.

\text{ControlNet \cite{zhang2023adding}} is included as a pose-guided diffusion baseline that conditions image synthesis using external structural guidance obtained from hand keypoint detectors. In our implementation, hand keypoints are extracted using MediaPipe whenever valid detections are available. However, these detectors are primarily designed for isolated hands and frequently fail under severe self-occlusions and hand-hand interactions commonly observed in Samyukta Hastas. Consequently, ControlNet provides a meaningful reference for pose-guided image synthesis while highlighting the challenges of applying generic keypoint-based conditioning to interacting-hand gesture generation.

\begin{figure*}
\centering
\includegraphics[width=\linewidth]{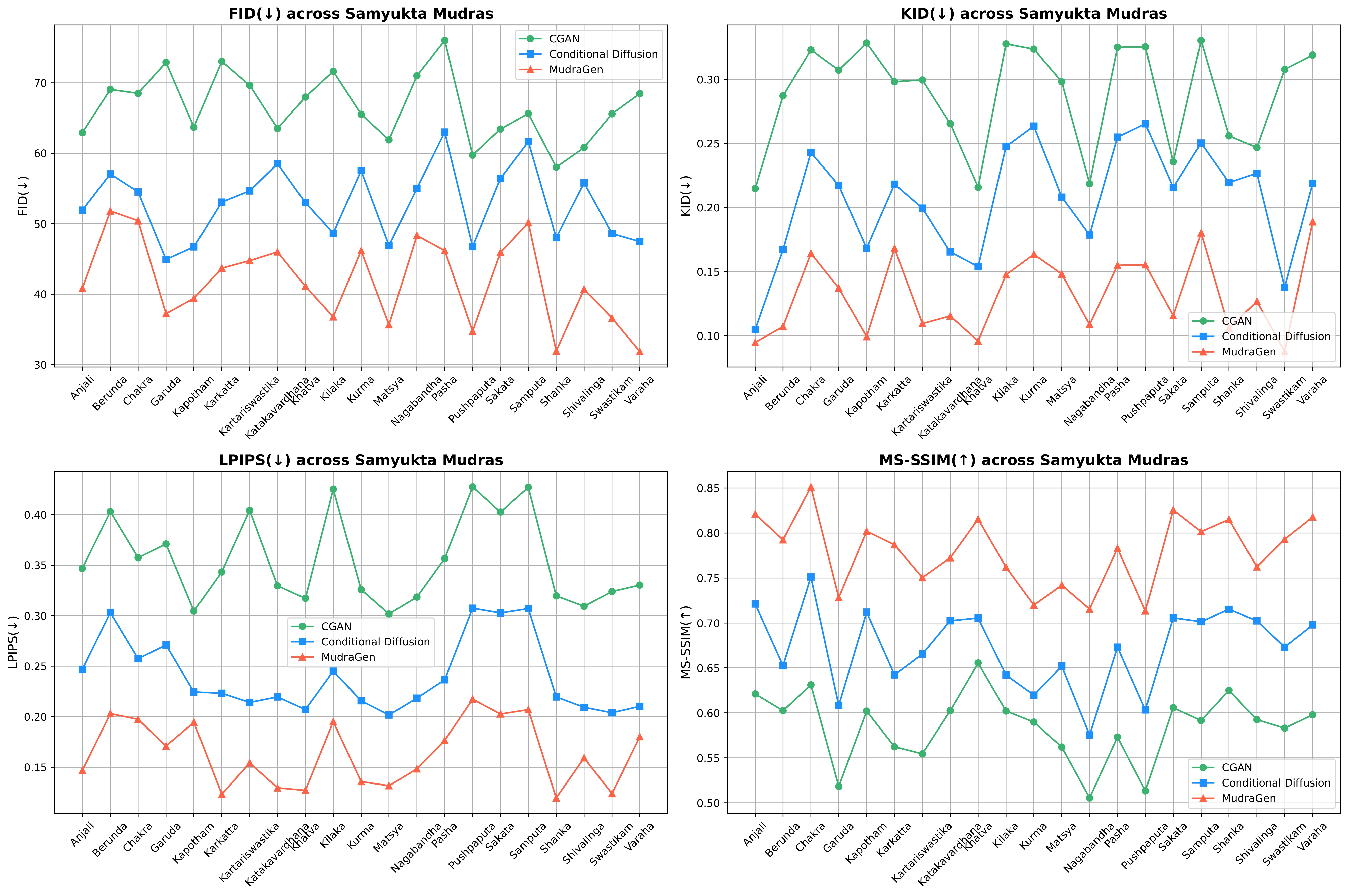}
\caption{    Class-wise quantitative comparison of generative models (Conditional GAN, Conditional Diffusion, MudraGen) across four metrics: FID , KID , LPIPS , and MS-SSIM. Lower FID, KID, and LPIPS values indicate higher realism and perceptual similarity, while higher MS-SSIM reflects better structural consistency. Our model consistently outperforms baselines across most gesture classes, particularly for complex two-hand configurations.}
\label{fig:metrics-comparison}
\end{figure*}

\paragraph{Quantitative Assessment:}
We use a set of standard evaluation metrics to quantitatively assess the fidelity and realism of the generated Samyukta Hasta Mudra images: Fréchet Inception Distance (FID) \cite{heusel2017gans}, Kernel Inception Distance (KID)\cite{binkowski2018demystifying}, LPIPS \cite{zhang2018unreasonable}, and MS-SSIM\cite{wang2003multiscale}. For quantitative evaluation, we fine-tune the Inception network on the curated Samyukta Hasta Mudra dataset to obtain feature representations that better capture the semantic and structural characteristics of interacting Bharatanatyam hand gestures. This domain-adapted feature extractor enables a more meaningful assessment of generative quality than a generic ImageNet-pretrained model. To ensure a fair and consistent comparison, the same fine-tuned Inception network is used uniformly for FID computation across all evaluated methods, including Stable Diffusion ~\cite{rombach2022high}, Hand1000 \cite{zhang2024hand1000}, Controlnet\cite{zhang2023adding}, and the proposed framework. Consequently, all models are assessed within an identical feature space, eliminating potential bias arising from differences in feature extraction. The quantitative results for all 21 Samyukta Mudra classes are summarized in Table~\ref{tab:overall-comparison}. FID and KID check how closely the generated images match the real data distribution in a deep feature space. They look at overall visual fidelity and semantic alignment. LPIPS measures perceptual similarity at the local level, which is important for evaluating fine-grained hand poses because it can pick up on small differences in how fingers move and how gestures look. MS-SSIM checks for global structural similarity, making sure that both hands are aligned and can interact with each other in a consistent way. These metrics give a full picture of how realistic and structurally sound the generated gestures are. Our model consistently surpasses current state-of-the-art models, attaining reduced FID, KID, and LPIPS values, alongside elevated MS-SSIM, signifying enhanced perceptual fidelity, realism, and structural coherence. To further analyze the generative quality across different Samyukta Mudra categories, we present class-wise performances utilizing the aforementioned four standard metrics. Figure \ref{fig:metrics-comparison} shows that our proposed model consistently does better than the baseline methods cGAN and Conditional Diffusion for gesture classes. The trends we saw in all four metrics show that our geometric supervision works to make both photorealism and anatomical plausibility possible. This breakdown shows how each model deals with different levels of gesture complexity, giving us more detailed information than just overall averages.

\begin{table}[t]
\centering
\caption{Comparative performance of generative models on Samyukta Hasta Mudra dataset.}
\label{tab:overall-comparison}
\begin{tabular}{lcccc}
\toprule
\textbf{Model} & \textbf{FID($\downarrow$)} & \textbf{KID($\downarrow$)} & \textbf{LPIPS ($\downarrow$)} & \textbf{MS-SSIM($\uparrow$)}  \\ 
\midrule
cGAN \cite{mirza2014conditional}& 66.6238 & 0.0288 & 0.3545 & 0.5853 \\ \hline
cDiffusion \cite{ho2022classifierfree}&  56.8669 & 0.0205 & 0.2402 & 0.7085 \\ \hline
  Stable Diffusion ~\cite{rombach2022high} & 60.4629 & 0.0257 &  0.2732 & 0.6743\\ \hline
  Hand1000 \cite{zhang2024hand1000} & 59.5483 & 0.0233 & 0.1920 & 0.7023\\ \hline
  {Controlnet\cite{zhang2023adding}} & 56.7624 & 0.0202 & 0.1703 & 0.7297\\ \hline
\textbf{MudraGen}  & \textbf{51.9126} & \textbf{0.0132}& \textbf{0.1640} & \textbf{0.7534} \\

\bottomrule
\end{tabular}
\end{table}

\paragraph{Ablation-Based Insights:}
We perform a series of ablation studies to find out how much each geometric loss component contributed to the overall loss. During training, selectively disabling the keypoint loss, joint offset loss, and Shape Consistency regularizer, the results are shown in Table \ref{tab:ablation-loss}. Our experiments demonstrate that each of these components fulfills a unique and synergistic function in enhancing the accuracy of generated two-hand gestures. Using all three losses together always gets better structural accuracy and semantic coherence. This shows how important they all are in helping the diffusion model make plausible and culturally valid samyukta hasta mudras.

\begin{table*}[t]
\centering
\caption{ Ablation study on the impact of geometric loss components in MudraGen. Shape consistency improves FID and MS-SSIM the most but other losses provide complimentary information to improve KID and LPIPS.}
\label{tab:ablation}
\begin{tabular}{@{}lcccc@{}}
\toprule
\textbf{Model} & \textbf{FID ($\downarrow$)} & \textbf{KID ($\downarrow$)} & \textbf{LPIPS ($\downarrow$)} & \textbf{MS-SSIM ($\uparrow$)}   \\ 
\midrule
Keypoint Loss ($\mathcal{L}_\text{kp}$)  & 56.2375 & 0.0194 & 0.2313 & 0.7143 \\ \hline
Joint Offset Loss ($\mathcal{L}_\text{offset}$) & 55.0842 & 0.0195 & 0.2287 & 0.7087 \\ \hline
Shape Consistency regularizer ($\mathcal{L}_\text{shape}$) & 52.1543 & 0.0203 & 0.2353 & 0.7236 \\ \hline
$\mathcal{L}_\text{kp}$ + $\mathcal{L}_\text{offset}$ & 59.0767 & 0.0151 & 0.2296 & 0.7459 \\ \hline
$\mathcal{L}_\text{kp}$ + $\mathcal{L}_\text{shape}$  & 53.7512 & 0.0143 & 0.2164 & 0.7453 \\ \hline
$\mathcal{L}_\text{offset}$ + $\mathcal{L}_\text{shape}$ & 52.5439 & 0.0149 & 0.1805 & 0.7460 \\\hline
$\mathcal{L}_\text{kp}$ + $\mathcal{L}_\text{offset}$ + $\mathcal{L}_\text{shape}$ & \textbf{51.9126} & \textbf{0.0132} & \textbf{0.1640} & \textbf{0.7534} \\ 
\bottomrule
\end{tabular}
\label{tab:ablation-loss}
\end{table*}

\begin{figure*}[t]
\centering
\includegraphics[width=\linewidth]{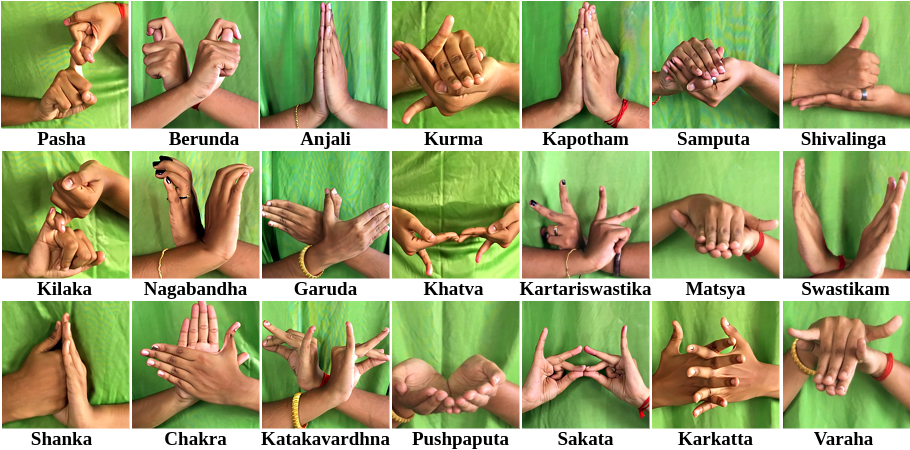}
\caption{Class-conditioned generation of all 21 Samyukta hasta mudra (two-hand gestures). Each column shows a generated image (128~$\times$~128 resolution) from a distinct gesture class, synthesized solely from Samyukta hasta mudra label. The samples exhibit anatomically consistent hand shapes, realistic left–right interactions, and semantically accurate finger configurations, demonstrating the model's ability to learn strong class-conditioned priors without relying on auxiliary inputs. }
\label{fig:qualitative}
\end{figure*}

\begin{figure*}
\centering
\includegraphics[width=\linewidth]{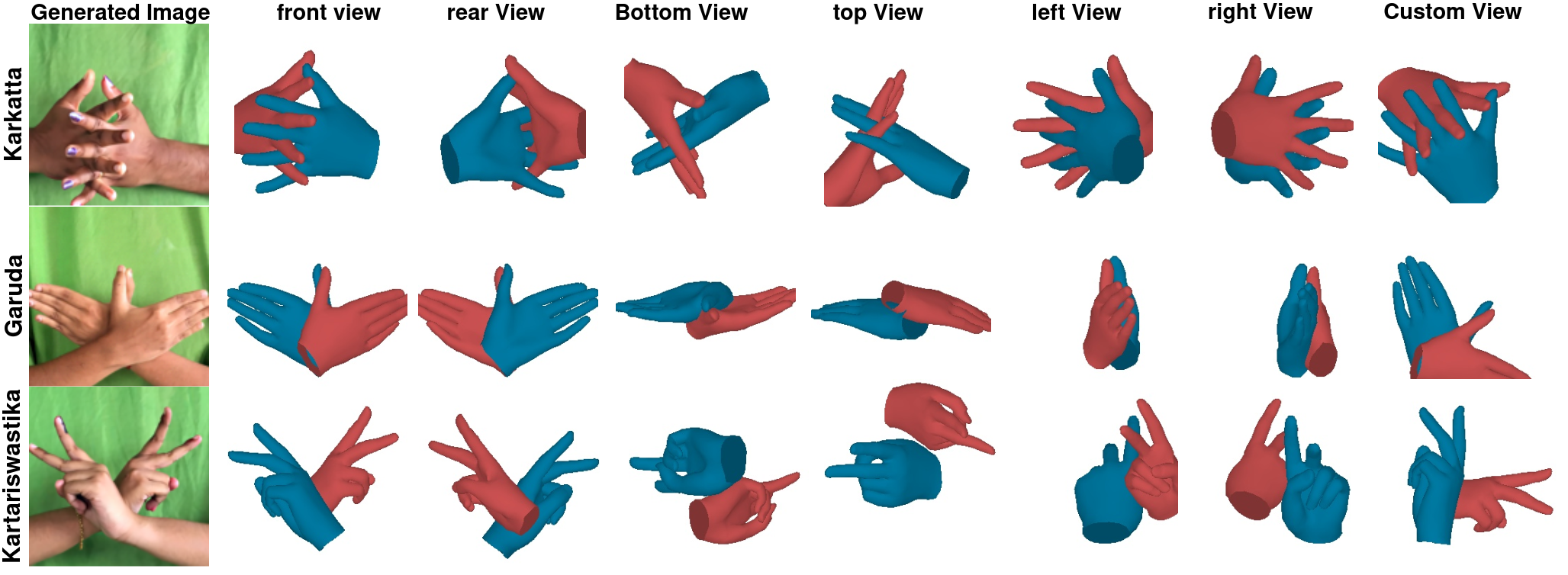}
\caption{Multi-view visualization of reconstructed 3D hand gestures from generated images. Each row begins with a synthesized RGB image, followed by seven rendered views of the corresponding 3D mesh: front, rear, left, right, top, bottom, and a custom oblique view. This layout facilitates qualitative evaluation of the geometric accuracy, articulation, and realism of the generated two-hand gestures from diverse viewpoints.}
\label{fig:multi_view_rendering}
\end{figure*}

\noindent
\begin{figure}[h]
  \centering
   \includegraphics[width=\linewidth]{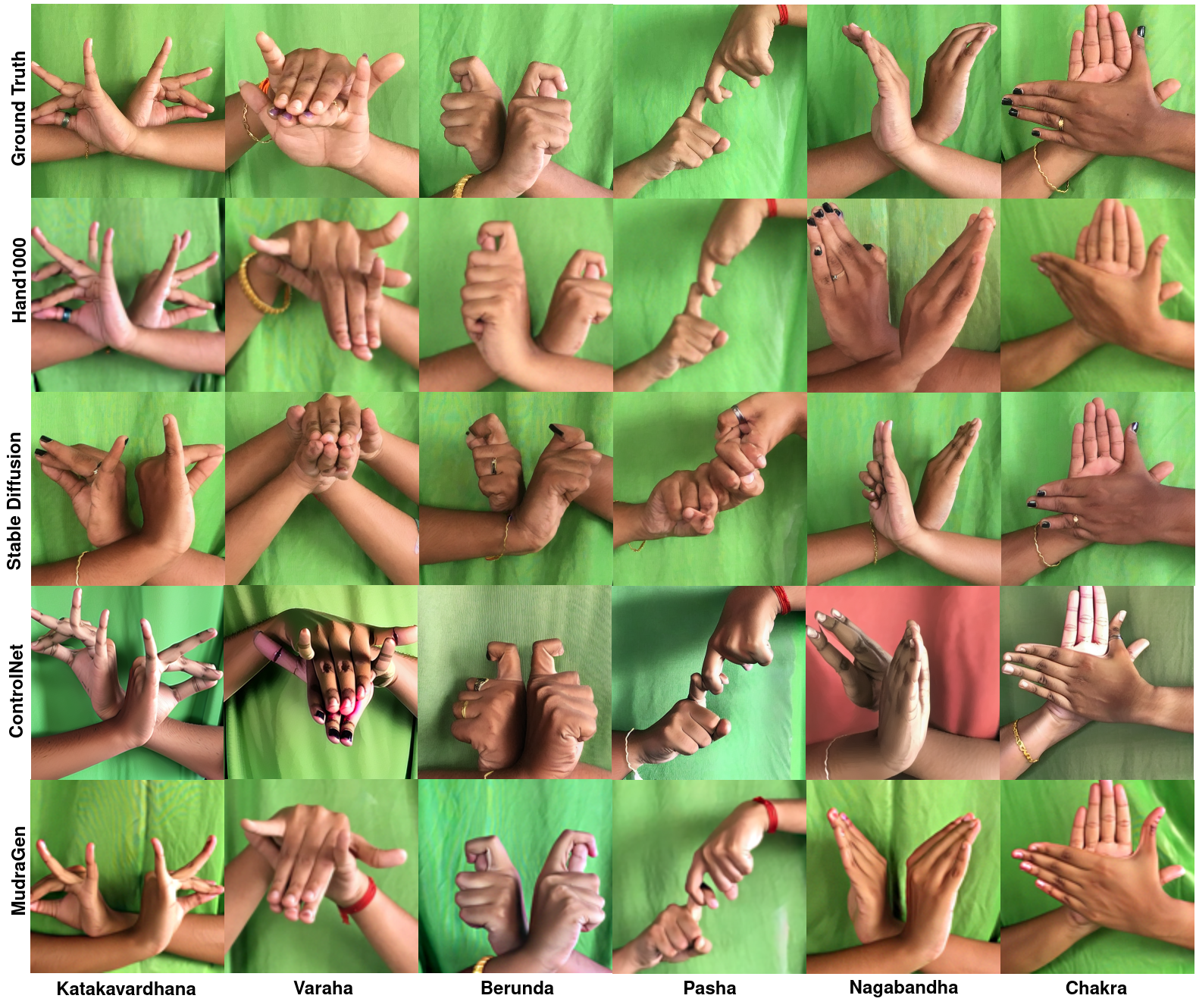}
   \caption{Qualitative results comparing ground truth (top row) with Hand1000 \cite{zhang2024hand1000}, Stable Diffusion \cite{rombach2022high}, {Controlnet\cite{zhang2023adding}} and our MudraGen model. Baseline models introduce geometric artifacts and implausible finger articulation, while our approach produces anatomically precise Samyukta Hastas with faithful gesture structure.}
   \label{fig:compare_vis_res}
\end{figure}
\vspace{-0.3cm}


\paragraph{Human Expert Evaluation of Generated Mudras}
Quantitative image-generation metrics cannot adequately measure the cultural correctness and semantic validity of Bharatanatyam hand gestures, we conducted a human evaluation study involving involving 24 practitioners and experts from Indian classical dance, each with more than eight to ten years of formal training. Participants were presented with generated images corresponding to the 21 Samyukta Hasta Mudras and were asked to identify the depicted mudra and evaluate its quality. As summarized in Table~\ref{tab:human_eval}, the assessment focused on key aspects essential to dance pedagogy and performance practice, namely gesture recognition, correctness of mudra formation, finger positioning accuracy, hand coordination, cultural authenticity, visual realism, and overall quality. All evaluations were collected using a five-point Likert scale. To minimize bias, participants were not informed that the images were generated by our system during the evaluation process. The blind evaluation setting ensured that judgments were based solely on the perceived quality and authenticity of the mudras. Following completion of the recognition task, participants were informed that the images had been synthesized by the proposed Bharatanatyam generation framework. They were subsequently asked to assess the potential of the system for educational use, digital preservation, and archival documentation of traditional dance knowledge.

The human evaluation serves two complementary objectives: (i) validating whether the generated mudras are recognizable and culturally faithful to established Bharatanatyam conventions, and (ii) assessing the practical utility of the proposed framework as a tool for dance education, cultural preservation, and long-term digital archiving.

\begin{table}[t]
\centering
\caption{ {Human evaluation results of generated Samyukta Hasta Mudras. Recognition accuracy is reported as a percentage, while all other metrics are measured on a five-point Likert scale (higher is better).}}
\small

\begin{tabular}{l c | l c}
\toprule
\multicolumn{2}{c|}{\textbf{Recognition \& Quality}} &
\multicolumn{2}{c}{\textbf{Educational \& Preservation}} \\
\midrule

Recognition Accuracy (\%) & 93.4 &
Cultural Faithfulness & 3.98 \\

Identification Confidence & 4.57 &
Educational Utility  & 4.12 \\

Mudra Formation & 4.23 &
Preservation Potential  &  3.90\\

Finger Accuracy & 4.15 & \\

Hand Coordination & 4.08 &
& \\

Cultural Authenticity & 4.35 &
& \\

Visual Realism & 4.31 &
& \\

Gesture Clarity & 4.46 &
& \\

Overall Quality & 4.12 &
& \\

\bottomrule
\end{tabular}
\label{tab:human_eval}
\end{table}

\vspace{-0.3cm}

\paragraph{Visual Analysis:}
Figure \ref{fig:qualitative} displays instances of generated samples from various gesture categories. We observe that the model can create realistic two-hand interactions with realistic poses and left-right hand alignment, even for gestures that require complicated finger movements and occlusions between hands. Even though they are only based on class labels, the samples have consistent anatomical structure, finger articulation, geometric realism, and hand symmetry. This shows that the model has learned a strong class that is conditioned on geometry. Our method works well for both simple gestures like "Anjali" and "Kapotham" and more complicated two-hand setups like "Kilaka" and "Pasha". Notably, the gap between methods becomes more pronounced in high-occlusion or contact-heavy poses, underscoring the importance of our geometric supervision strategy for anatomically plausible generation. MudraGen gets better structural and anatomical fidelity by using geometry-aware diffusion supervision. Using Keypoint alignment, inter-hand joint offset, and shape consistency constraints together improves the accuracy of articulated poses, makes sure that it enforces spatial coherence between hand, and keeps symmetry of the body. We use Fig.\ref{fig:multi_view_rendering} to show the reconstructed 3D hand meshes from different standard angles so that we can qualitatively assess the anatomical coherence and articulation of the gesture we made. These renderings from different angles help to prove that inter-hand interactions are possible and the structure is strong from all angles. This kind of visualization also makes it easier for novice and domain experts like classical dancers, to understand by letting them look at the gesture fidelity from different angles. MudraGen consistently outperforms state-of-the-art-generative-models, which often show distorted finger articulations and implausible hand shapes, especially in complex two-hand mudras, while still keeping semantic and visual realism, as shown in Fig.\ref{fig:compare_vis_res}.

\vspace{-0.3cm}

\subsection{Discussion}

The experimental results demonstrate that incorporating geometry-aware supervision into the diffusion learning process substantially improves the structural fidelity of generated Samyukta Hasta Mudras. Unlike conventional image generation approaches that primarily optimize visual appearance, the proposed framework explicitly encourages anatomically plausible finger articulation, inter-hand coordination, and consistent interacting-hand geometry. Consequently, the generated gestures preserve the semantic characteristics that distinguish visually similar Samyukta Hasta categories while maintaining realistic hand morphology. The improvements observed in both quantitative metrics and qualitative comparisons indicate that integrating geometric priors effectively addresses the challenges posed by low-resource Bharatanatyam datasets, where learning anatomically consistent interacting-hand representations from RGB images alone is particularly difficult.

The comparative evaluation further demonstrates the importance of incorporating explicit geometric information during optimization. Although general-purpose diffusion models and hand-specific image generation frameworks are capable of synthesizing visually realistic hand images, they often struggle to preserve the fine-grained structural relationships required for culturally meaningful interacting-hand gestures. By introducing pose, joint, and shape-based geometric regularization during training, the proposed framework consistently generates anatomically coherent Samyukta Hastas while preserving the coordinated spatial configuration between both hands.

Although this work focuses on Bharatanatyam Samyukta Hastas, the proposed geometry-aware framework is not inherently restricted to a specific cultural domain. By learning general geometric priors for coordinated interacting-hand articulation rather than relying solely on domain-specific appearance cues, the model captures structural relationships that are transferable across gesture-centric tasks. Consequently, the proposed methodology can be adapted to other low-resource interacting-hand applications, including sign language generation, traditional dance forms, gesture-based human–computer interaction, and other scenarios involving complex bimanual hand interactions.

Beyond image synthesis, the proposed framework offers practical value for cultural heritage preservation and dance education. The generated Samyukta Hasta images preserve essential structural characteristics, including finger articulation, inter-hand coordination, symmetry and asymmetry patterns, and canonical gesture geometry. Furthermore, the generated images can be processed using a pretrained 3D hand reconstruction network to obtain multi-view visualizations without requiring motion-capture systems or ground-truth 3D annotations. These representations enable learners and practitioners to examine complex finger configurations from multiple viewpoints, including regions affected by self-occlusion, thereby supporting digital archiving, remote learning, interactive educational platforms, and long-term preservation of Indian classical dance traditions.

\subsection{Limitations}

Although the proposed framework demonstrates promising results for anatomically consistent Samyukta Hasta Mudra generation, some limitations remain that provide opportunities for future research.
Although the proposed geometry-aware supervision enables effective learning under limited data availability, the diversity of performer appearances, viewpoints, and environmental conditions remains considerably smaller than that of large-scale image generation datasets. 
Another limitation of the current framework is that it synthesizes a single-view RGB image for a given Samyukta Hasta Mudra, while additional viewpoints are obtained through post-generation 3D hand reconstruction. Although these reconstructed meshes enable multi-view visualization, they do not directly provide photo-realistic RGB appearances from arbitrary viewpoints. 

\section{Conclusion}

\label{sec:conclusion}
In this work, we presented a novel diffusion-based framework for generating culturally significant two-hand gestures (Samyukta Hasta Mudras) directly from discrete class labels. Our method uniquely combines semantic conditioning via class embeddings with geometric supervision derived from a pretrained 3D hand mesh estimator. By supervising the generated images with keypoint alignment, joint offset consistency, and shape symmetry losses, we ensure anatomical plausibility and gesture correctness, even in complex, interacting hand poses. In contrast to prior methods that depend on rich multi-modal inputs like 3D meshes or 2D Keypoints, our approach learns to synthesize hand gesture images conditioned only on class labels during inference—making it lightweight, efficient, and more suitable for real-world deployment. Additionally, we demonstrate how the generated images can be further interpreted by reconstructing 3D hand meshes and rendering multi-view visualizations, offering a complete loop from label to geometry. This research contributes both a computational approach and a cultural preservation tool, enabling the generative synthesis of symbolically rich gestures rooted in Indian classical traditions. Future research may explore integrating temporal modeling for gesture sequences and expanding to other forms of cultural expression.

\bibliographystyle{plain}
\bibliography{references}  






\end{document}